# A Fuzzy Petri Nets Model for Computing With Words

Yongzhi Cao and Guoqing Chen

*Abstract*— Motivated by Zadeh's paradigm of computing with words rather than numbers, several formal models of computing with words have recently been proposed. These models are based on automata and thus are not well-suited for concurrent computing. In this paper, we incorporate the well-known model of concurrent computing, Petri nets, together with fuzzy set theory and thereby establish a concurrency model of computing with words—fuzzy Petri nets for computing with words (FPNCWs). The new feature of such fuzzy Petri nets is that the labels of transitions are some special words modeled by fuzzy sets. By employing the methodology of fuzzy reasoning, we give a faithful extension of an FPNCW which makes it possible for computing with more words. The language expressiveness of the two formal models of computing with words, fuzzy automata for computing with words and FPNCWs, is compared as well. A few small examples are provided to illustrate the theoretical development.

## I. INTRODUCTION

IT is well known that numerical precision is an expensive and often unnecessary goal in describing the world around us. On the other hand, human thought does not operate at a numeric level of precision, but at a much more abstract level. To formally express that abstraction within computational processes, Zadeh has proposed and advocated the idea of computing with words (CW, also known as granular computing) in a series of papers [46]–[50]. CW aims at capturing the automated reasoning involving linguistic terms rather than numerical quantities. Clearly, such a reasoning is of central importance for endowing computer systems with a more human-centric view of the world. Ever since the introduction of the term of CW, we have witnessed a rapid development of and a fast growing interest in the topic (see, for example, [3], [6], [11], [12], [17]–[19], [22], [34], [38]–[44], [51]).

In most of the literature on CW, the word "computing" in the phrase "computing with words" refers only to computationally efficient mechanisms for modeling and reasoning under uncertain conditions, not to any formal theory of computing. However, computing, in its traditional sense, is centered on

This work was supported by the National Natural Science Foundation of China under Grants 70890080 and 60973004 and by the National Basic Research Program of China (973 Program) under Grants 2007CB311003 and 2009CB320701.

Y. Cao is with the Institute of Software, School of Electronics Engineering and Computer Science, Peking University, Beijing 100871, China, and the Key Laboratory of High Confidence Software Technologies (Peking University) Ministry of Education, China (e-mail: caoyz@pku.edu.cn).

G. Chen is with the School of Economics and Management, Tsinghua University, Beijing 100084, China (e-mail: chengq@sem.tsinghua.edu.cn).

manipulation of numbers or symbols, and is usually represented by a dynamic model such as all kinds of automata and Petri nets. Since classical models of computation aim at describing numerical or symbolical calculation, their inputs are usually supposed to be exact rather than vague data. Motivated by this observation and Zadeh's paradigm of CW, Ying [44] put forward two kinds of fuzzy automata, which accept fuzzy inputs, as formal models of CW. More specifically, he modeled the words in the CW paradigm by fuzzy subsets of a set of symbols and took all these words as the input alphabet of such fuzzy automata. Instead of accepting or rejecting a string of words, these fuzzy automata will accept the string with a certain degree between zero and one. Such an idea has been developed for probabilistic automata and fuzzy Turing machines in [34] and [39], respectively.

Recently, the authors and Ying have established a general formal model of computing with (some special) words via fuzzy automata in [6]. The new features of the model are that the input alphabet only comprises some (not necessarily all) words modeled by fuzzy subsets of a set of symbols and the fuzzy transition function can be specified arbitrarily. By employing the methodology of fuzzy control, we have obtained a retraction principle from CW to computing with values for handling crisp inputs and a generalized extension principle from CW to computing with all words for handling fuzzy inputs.

It is worth noting that all the formal models of CW mentioned above are based upon automata. These automata are the simplest computational models which have the advantages of being intuitive, amenable to composition operations, and amenable to analysis as well. On the other hand, it is well known that automata are not satisfactory for describing concurrent computing (see, for example, [27]), and moreover, they lack structure and for this reason may lead to very large state spaces when modeling some complex computations. An alternative to automata for formal models of computing is provided by Petri nets. Petri Nets, first developed by C. A. Petri in the early 1960's [32], are a formal and graphical appealing language which is appropriate for modeling systems with concurrency and resource sharing [28], [31]. Petri net models have more structure than automaton models, although they do not possess, in general, the same analytical power as automata.

The purpose of this paper is to develop a concurrency model of CW by exploiting Petri nets. Since impreciseness and uncertainty are more or less involved in CW, we would like to take fuzzy Petri nets (FPNs), which combine fuzzy set theory and Petri net theory, as a computational model of CW.

In the literature, there are various approaches for combining Petri nets and fuzzy sets (see [1], [2], [7]–[10], [15], [20], [29], [36] and the references therein); they all are called fuzzy Petri nets. Under the same name, these models are based on different notions. They fall roughly into two styles: either the FPN represents a dynamic system and the marking denotes the uncertain information about its current state or it depicts a chaining of some fuzzy reasoning rules and the marking corresponds to some step within a reasoning. To the best of our knowledge, few efforts are made to consider the fuzziness of transitions and their labels, although most of the works have a certainty or confidence factor or a threshold value associated with each transition. Following the interpretation of CW in [44], it is natural to allow the labels to be words (i.e., fuzzy subsets of a finite set of symbols), since the labels of transitions exactly correspond to the input alphabet of an automaton. As we will see later, allowing the labels to represent vague data is sometimes beneficial for modeling and reasoning, as we will see from the subsequent examples. This observation motivates us to adopt FPNs with word labels, which we will refer to as fuzzy Petri nets for computing with words (FPNCWs), as a formal model of CW. Our FPNCWs inherit some good properties of the ordinary Petri nets; in particular, the concept of concurrently occurring events can be expressed directly. In addition, the FPNCW model has certain advantages in some compositions of sub-models.

In practice, taking time, cost, and other factors into account one may establish only an FPNCW for some special words. To make the model possible for computing with more words, we embark upon an extension by using fuzzy reasoning. The starting point of the extension is an FPNCW modeling computing with (some special) words. The resultant model is still an FPN, but it has more transitions and labels (words) than the original FPNCW, and thus it is called a fuzzy Petri net for computing with more words (FPNCMW). Unlike the generalized extension in [6], the extension here is faithful in the sense that for those words appearing in the original FPNCW, both the original and extended models yield the same calculations. In fact, the extension provides an interpolation approach which helps reduce the complexity of devising a model for CW. This FPNCMW, which serves as a CW engine, may accept some strings of words as inputs and give corresponding acceptable degrees (or final states) as outputs. Therefore, our CW engine is distinctly different from the CW engines appearing in the literature such as Mendel's perceptual reasoning [24]–[26], IF-THEN rules, linguistic summarizations, and linguistic weighted averages; these CW engines map their input words into their output words.

The remainder of this paper is structured as follows. In Section II, after briefly recalling some basics of fuzzy set theory, we introduce the concurrency model of CW. Section III is devoted to the extension from computing with words to computing with more words, including building a rule base, developing a reasoning algorithm, and investigating the formal languages represented by the extended model. We compare the language expressiveness of the two formal models of CW, fuzzy automata for CW and FPNCWs, in Section IV and conclude the paper in Section V. The proofs of our theorems and propositions are given in Appendix A.

## II. CONCURRENCY MODEL OF COMPUTING WITH WORDS

To introduce a Petri net model of CW, let us first review some notions on fuzzy set theory and then tailor an FPN according to our purpose.

### A. Fuzzy Sets

Let $X$ be a universal set. A *fuzzy set* $A$ [45], or rather a *fuzzy subset* $A$ of $X$, is defined by a function assigning to each element $x$ of $X$ a value $A(x)$ in the closed unit interval $[0, 1]$. Denote by $\mathcal{F}(X)$ the set of all fuzzy subsets of $X$. For any $A, B \in \mathcal{F}(X)$, we say that $A$ is contained in $B$ (or $B$ contains $A$), denoted by $A \subseteq B$, if $A(x) \leq B(x)$ for all $x \in X$. We say that $A = B$ if and only if $A \subseteq B$ and $B \subseteq A$. A fuzzy set is said to be *empty* if its membership function is identically zero on $X$. We use $0$ to denote the empty fuzzy set.

The *support* of a fuzzy set $A$ is a crisp set defined as $\mathrm{supp}(A) = \{x \in X : A(x) > 0\}$. Whenever $\mathrm{supp}(A)$ is finite, say $\mathrm{supp}(A) = \{x_1, x_2, \ldots, x_n\}$, we may write $A$ in Zadeh's notation as

$$A = \frac{A(x_1)}{x_1} + \frac{A(x_2)}{x_2} + \cdots + \frac{A(x_n)}{x_n}.$$

Given $A, B \in \mathcal{F}(X)$, the *union* of $A$ and $B$, denoted $A \cup B$, is defined by the membership function

$$(A \cup B)(x) = A(x) \vee B(x)$$

for all $x \in X$; the *intersection* of $A$ and $B$, denoted $A \cap B$, is given by the membership function

$$(A \cap B)(x) = A(x) \wedge B(x)$$

for all $x \in X$. Let $\lambda \in [0, 1]$ and $A \in \mathcal{F}(X)$. The *scale product* $\lambda \cdot A$ of $\lambda$ and $A$ is defined by

$$(\lambda \cdot A)(x) = \lambda \wedge A(x)$$

for every $x \in X$; this is again a fuzzy subset of $X$.

For any family $\lambda_i$, $i \in I$, of elements of $[0, 1]$, we write $\vee_{i \in I} \lambda_i$ or $\vee \{\lambda_i : i \in I\}$ for the supremum of $\{\lambda_i : i \in I\}$, and $\wedge_{i \in I} \lambda_i$ or $\wedge \{\lambda_i : i \in I\}$ for the infimum. In particular, if $I$ is finite, then $\vee_{i \in I} \lambda_i$ and $\wedge_{i \in I} \lambda_i$ are the greatest element and the least element of $\{\lambda_i : i \in I\}$, respectively. For any $A \in \mathcal{F}(X)$, the *height* of $A$ is defined as

$$\mathrm{height}(A) = \vee_{x \in X} A(x).$$

For a detailed introduction to the above notions, we refer the reader to [13], [30].

### B. Fuzzy Petri Nets for Computing With Words

As mentioned above, there are various approaches for combining Petri nets and fuzzy sets in the literature (see [1], [2], [7]–[10], [15], [20], [29], [36] and the references therein). Let us tailor the following FPN according to our purpose.

*Definition 1:* A *fuzzy Petri net* (or FPN for short) is defined as a seven-tuple

$$\mathcal{N} = (P, T, I, O, \alpha, \beta, M_0),$$

where
1) $P = \{p_1, p_2, \ldots, p_n\}$ is a finite set of places.
2) $T = \{t_1, t_2, \ldots, t_m\}$ is a finite set of transitions.
3) $I \subseteq P \times T$ is a set of directed arcs from places to transitions. We call each $p_i$ where $(p_i, t_j) \in I$ as an *input place* of $t_j$.
4) $O \subseteq T \times P$ is a set of directed arcs from transitions to places. We call each $p_i$ where $(t_j, p_i) \in O$ as an *output place* of $t_j$.
5) $\alpha : T \longrightarrow (0, 1]$ is an association function assigning to each transition a threshold value between zero and one.
6) $\beta : O \longrightarrow (0, 1]$ is an association function assigning to each directed arc from transitions to places, a fuzzy truth value between zero and one.
7) $M_0 : P \longrightarrow [0, 1]$ is an association function, representing the initial state of the net. For each $p_i \in P$, $M_0(p_i)$ indicates the degree to which $p_i$ is an initial state.

We assume that all FPNs under consideration have no isolated places or transitions. In describing an FPN, it is convenient to use $I(t_j)$ to represent the set of input places to transition $t_j$, namely, $I(t_j) = \{p_i : (p_i, t_j) \in I\}$. Similarly, $O(t_j)$ represents the set of output places from transition $t_j$, namely, $O(t_j) = \{p_i : (t_j, p_i) \in O\}$. Similar notation can be used to describe input and output transitions for a given place $p_i$: $I(p_i)$ and $O(p_i)$.

When drawing fuzzy Petri net graphs, we need to differentiate between the two types of nodes, places and transitions. Following the convention, we use circles to represent places and bars to represent transitions. The arcs directed from places to transitions represent elements of $I$, and the arcs directed from transitions to places represent elements of $O$. For any $t_j \in T$ with $\alpha(t_j) = c_j$, we attach $c_j$ with the bar that represents the transition $t_j$. Similarly, if $(t_j, p_i) \in O$ with $\beta(t_j, p_i) = w_{ji}$, we label the directed arc from $t_j$ to $p_i$ with $w_{ji}$.

Whenever $M_0(p_i) > 0$, we assign a token with label $M_0(p_i)$ to the place $p_i$. Such a labeled token is something we "put in a place" essentially to indicate the fact that the condition described by the place $p_i$ is satisfied with the possibility degree $M_0(p_i)$. More generally, the way in which tokens are assigned to a fuzzy Petri net graph defines a *marking*. Formally, a *marking* $M$ is a function $M : P \longrightarrow [0, 1]$. Thus, $M_0$ is a marking, called the *initial marking*. Notice that a marking $M$ defines a row vector $M = [M(p_1), M(p_2), \ldots, M(p_n)] \in [0, 1]^n$, where $n$ is the number of places in the FPN. The marking row vector is also referred to as the *state* of the FPN. In fuzzy Petri net graphs, a token is indicated by a dark dot positioned in the appropriate place.

For the sake of illustrating the above notion and notation, let us see a simple example.

*Example 1:* Consider a mixed water valve that accommodates the temperature of water from a hot water pipe and a cold water pipe. The water valve has typically three states: *almost fully open hot water*, *almost fully open cold water*, and *half open hot (cold) water*. The water temperature of outflow is experientially classified as three states: $high$, $medium$, and $low$. Because the classification is imprecise, we are not certain about the exact state of the water temperature of outflow.

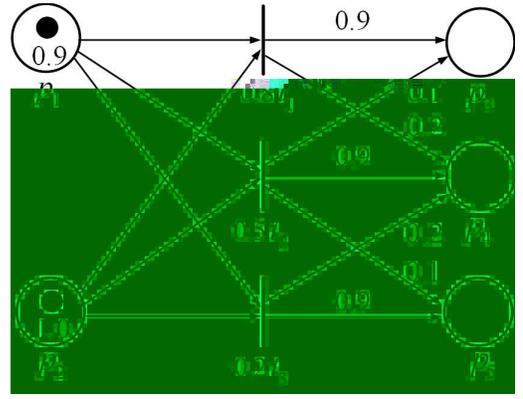

Fig. 1. An FPN modeling the relationships among the water temperature of inflow, the water valve, and the water temperature of outflow.

Consider the case that the water from the hot water pipe is certainly hot and the water from the cold water pipe is certainly cold. Let us suppose, for example, that in this case, the water temperature of outflow is $high$ with the possibility degree $0.9$ and $medium$ with the possibility degree $0.2$ if the water valve is in the state of *almost fully open hot water*; the water temperature of outflow is $low$ with the possibility degree $0.9$ and $medium$ with the possibility degree $0.2$ if the water valve is in the state of *almost fully open cold water*; the water temperature of outflow is $high$ with the possibility degree $0.1$, $medium$ with the possibility degree $0.9$, and $low$ with the possibility degree $0.1$ if the water valve is in the state of *half open hot (cold) water*.

We would like to model the relationships among the water temperature of inflow, the water valve, and the water temperature of outflow by an FPN (see Fig. 1). Formally, we specify the places and transitions of the FPN as follows.

$p_1$: "the water from the hot water pipe is $hot$";
$p_2$: "the water from the cold water pipe is $cold$";
$p_3$: "the water temperature of outflow is $high$";
$p_4$: "the water temperature of outflow is $medium$";
$p_5$: "the water temperature of outflow is $low$";
$t_1$: "the water valve is in the state of *almost fully open hot water*";
$t_2$: "the water valve is in the state of *half open hot (cold) water*";
$t_3$: "the water valve is in the state of *almost fully open cold water*."

According to the previous arguments, we have that
$I = \{(p_1, t_1), (p_1, t_2), (p_1, t_3), (p_2, t_1), (p_2, t_2), (p_2, t_3)\}$;
$O = \{(t_1, p_3), (t_1, p_4), (t_2, p_3), (t_2, p_4), (t_2, p_5), (t_3, p_4), (t_3, p_5)\}$;
$\beta(t_1, p_3) = \beta(t_2, p_4) = \beta(t_3, p_5) = 0.9$, $\beta(t_1, p_4) = \beta(t_3, p_4) = 0.2$, $\beta(t_2, p_3) = \beta(t_2, p_5) = 0.1$.

Assume that the threshold values assigning to the transitions $t_1$, $t_2$, and $t_3$ are $0.8$, $0.5$, and $0.2$, respectively, namely, $\alpha(t_1) = 0.8$, $\alpha(t_2) = 0.5$, and $\alpha(t_3) = 0.2$. We also assume that the initial marking $M_0$ is $[0.9, 1, 0, 0, 0]$, which means that the water from the hot water pipe is $hot$ with possibility degree $0.9$ and the water from the cold water pipe is $cold$ with possibility degree $1$.

$$M'(p_k) = \begin{cases} M(p_k) \vee (\mu_{M,t_j} \wedge \beta(t_j, p_k)), & \text{if } p_k \notin I(t_j) \\ \mu_{M,t_j} \wedge \beta(t_j, p_k), & \text{if } p_k \in I(t_j) \bigcap O(t_j) \\ 0, & \text{if } p_k \in I(t_j) \backslash O(t_j) \end{cases}$$

We are now in the position to describe the state transition mechanism of FPNs. To this end, we first need to introduce the notion of *enabled transition*.

*Definition 2:* A transition $t_j \in T$ in an FPN is said to be *enabled* if $M(p_i) \geq \alpha(t_j)$ for every $p_i \in I(t_j)$.

In other words, a transition $t_j$ in the FPN is enabled if and only if the possibility degree of $p_i$ is at least as large as the threshold value of $t_j$, for all places $p_i$ that are input to transition $t_j$. For later need, we write $\mu_{M,t_j}$ for $\wedge_{p_i \in I(t_j)} M(p_i)$. With this notation, we see that the transition $t_j$ is enabled if $\mu_{M,t_j} \geq \alpha(t_j)$.

The state transition mechanism in FPNs is provided by *moving tokens* with possibility degrees through the net and hence changing the state of the FPN. To explicitly present the current state of an FPN, we sometimes refer to $(P, T, I, O, \alpha, \beta, M)$ as an FPN, where $M$ stands for the current state. When a transition is enabled, we say that it can *fire*. The state transition function of an FPN is defined through the change in the state of the FPN due to the firing of an enabled transition. More precisely, we have the following definition.

*Definition 3:* The *state transition function*, $f : [0,1]^n \times T \longrightarrow [0,1]^n$, of an FPN $(P, T, I, O, \alpha, \beta, M)$ is defined for transition $t_j \in T$ if and only if $\mu_{M,t_j} \geq \alpha(t_j)$, that is, $M(p_i) \geq \alpha(t_j)$ for every $p_i \in I(t_j)$. Furthermore, if $f(M, t_j)$ is defined, then we set $M' = f(M, t_j)$, where for any $p_k \in P$, $M'(p_k)$ is given by the equation at the top of this page.

Therefore, after firing the transition $t_j$, if $p_k$ is an input place of $t_j$, it loses a token; if it is an output place of $t_j$, it gains a token. Clearly, it is possible that $p_k$ is both an input and output place of $t_j$, in which case the token is at first removed from $p_k$, and then a new token is immediately placed back in it. What we should keep in mind is that after the firing the label $M(p_k)$ must be replaced by $M'(p_k)$.

As an example, let us examine the state transition mechanism of the FPN in Example 1.

*Example 2:* Consider the FPN in Example 1. By definition we see that all the three transitions in the FPN are enabled. For instance, let us fire the transition $t_2$. Writing $M'$ for $f(M_0, t_2)$,

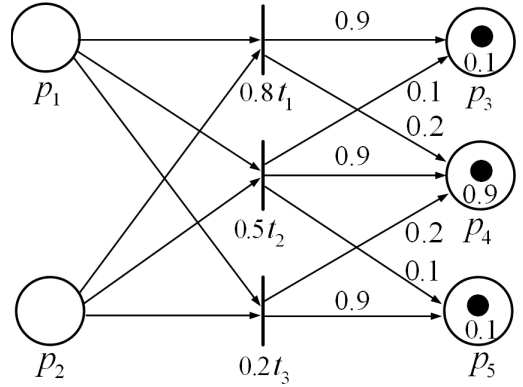

Fig. 2. Firing the transition $t_2$ of the FPN in Fig. 1.

we obtain immediately from definition that

$$\begin{aligned} M'(p_1) &= 0; \\ M'(p_2) &= 0; \\ M'(p_3) &= M_0(p_3) \vee (\mu_{M_0,t_2} \wedge \beta(t_2, p_3)) \\ &= 0 \vee [(0.9 \wedge 1.0) \wedge 0.1] \\ &= 0.1; \\ M'(p_4) &= M_0(p_4) \vee (\mu_{M_0,t_2} \wedge \beta(t_2, p_4)) \\ &= 0 \vee [(0.9 \wedge 1.0) \wedge 0.9] \\ &= 0.9; \\ M'(p_5) &= M_0(p_5) \vee (\mu_{M_0,t_2} \wedge \beta(t_2, p_5)) \\ &= 0 \vee [(0.9 \wedge 1.0) \wedge 0.1] \\ &= 0.1. \end{aligned}$$

As a result, $M' = [0, 0, 0.1, 0.9, 0.1]$. The fuzzy Petri net graph after firing the transition $t_2$ is depicted in Fig. 2.

To describe what happens when we fire a sequence of transitions, we need to extend the state transition function $f$ from domain $[0,1]^n \times T$ to $[0,1]^n \times T^*$:

$$\begin{aligned} f(M, \epsilon) &:= M \\ f(M, st) &:= f(f(M, s), t) \text{ for } s \in T^* \text{ and } t \in T, \end{aligned}$$

where the empty string $\epsilon$ is to be interpreted as the absence of transition firing and $T^*$ is the set of all finite strings over $T$, including the symbol $\epsilon$. In the literature of classical computation theory, a string is often called a "word". Like [44], to avoid confusion in this paper, we do not use the term "word" in this way and only use it to refer to what we mean by "word" in the phrase "computing with words."

For later need, let us introduce the concept of reachable states.

*Definition 4:* Let $\mathcal{N} = (P, T, I, O, \alpha, \beta, M_0)$ be an FPN. A marking $M \in [0,1]^n$ is *reachable* (from $M_0$) in $\mathcal{N}$ if there exists $s \in T^*$ such that $f(M_0, s) = M$. Denote by $R(\mathcal{N})$ the set of all reachable states in $\mathcal{N}$.

For example, it is easy to see that the FPN in Example 1 has four reachable states: $M_0$, $[0, 0, 0.9, 0.2, 0]$, $[0, 0, 0.1, 0.9, 0.1]$, and $[0, 0, 0, 0.2, 0.9]$.

The above definitions of the extended form of the state transition function and of the set of reachable states assume that enabled transitions fire *one at a time*. Clearly, some transitions that "consume" tokens from disjoint sets of places could fire simultaneously. Since we are interested in all possible states that can be reached, and since later on we will be labeling transitions with words and considering the languages accepted by FPNs, we shall henceforth exclude such simultaneous firings of transitions and assume that transitions fire one at a time.

For the aforementioned notions, we have two remarks.

*Remark 1:* Recall that in classical Petri net theory, a place is said to be 1-*safe* (or simply *safe*) if the number of tokens in that place cannot exceed 1, and a Petri net is called *safe* if every place of the net is safe. Clearly, Definitions 1-4 generalize those of safe Petri nets and reduce to safe Petri nets when the images of $\alpha$ and $\beta$ are $\{1\}$ and the image of $M_0$ is $\{0, 1\}$.

*Remark 2:* A careful reader may find that in Definition 1 the association function assigning fuzzy truth values to directed arcs is not symmetric, that is, values are only associated to directed arcs from transitions to places, but not to those from places to transitions. Indeed, similar to [36] we may add one more association function, say, $w : I \longrightarrow (0, 1]$ to the definition of FPNs for representing fuzzy truth value, inward flux, or something alike. It may be helpful to modeling in some practical questions, but theoretically it seems redundant since we can transform an FPN $\mathcal{N}_w$ with the association function $w$ into an FPN $\mathcal{N}$ in the sense of Definition 1 that has the same state transition mechanism as $\mathcal{N}_w$, and vice versa.

More concretely, let $\mathcal{N}_w = (P, T, I, O, \alpha, \beta, M_0, w)$, and following Definition 3 we may suppose that the state transition function $f_{\mathcal{N}_w}$ of $\mathcal{N}_w$ is defined for transition $t_j \in T$ if and only if $\mu_{M, t_j} \geq \alpha(t_j)$. Furthermore, if $f_{\mathcal{N}_w}(M, t_j)$ is defined, it is rational to take the value of $f_{\mathcal{N}_w}(M, t_j)(p_k)$ for any $p_k \in P$ as in the equation at the top of the page. Now, we may define an FPN $\mathcal{N} = (P, T, I, O, \alpha, \beta_w, M_0)$, where $\beta_w : O \longrightarrow (0, 1]$ is given by

$$\beta_w(t_j, p_i) = [\bigwedge_{p_k \in I(t_j)} w(p_k, t_j)] \wedge \beta(t_j, p_i)$$

for any $(t_j, p_i) \in O$. Then it is easy to see by Definition 3 that the state transition functions of $\mathcal{N}$ and $\mathcal{N}_w$ are exactly equal. Conversely, given an FPN $\mathcal{N} = (P, T, I, O, \alpha, \beta, M_0)$, we may take $w(p_i, t_j) = 1$ for each $(p_i, t_j) \in I$. As a result, we obtain an FPN $\mathcal{N}_w = (P, T, I, O, \alpha, \beta, M_0, w)$. Again, it is easy to see that the state transition function of $\mathcal{N}_w$ is the same as that of $\mathcal{N}$.

The transformation from $\mathcal{N}$ to $\mathcal{N}_w$ is trivial and there is nothing to explain. To illustrate the inverse transformation, let us see a simple example.

*Example 3:* Consider the FPN $\mathcal{N}_w = (P, T, I, O, \alpha, \beta, M_0, w)$ depicted in Fig. 3, where

- $P = \{p_1, p_2, \ldots, p_5\}$;
- $T = \{t_1, t_2, t_3\}$;
- $I = \{(p_1, t_1), (p_1, t_2), (p_1, t_3), (p_2, t_1), (p_2, t_2), (p_2, t_3)\}$;

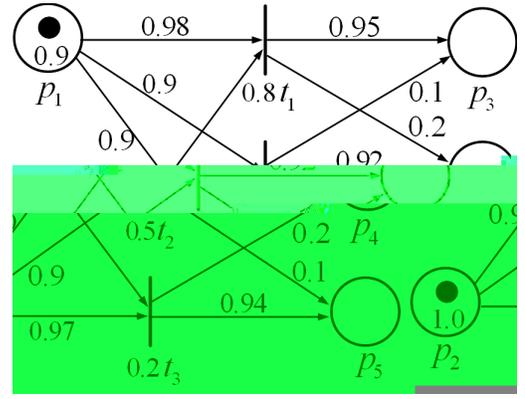

Fig. 3. An FPN with the association function assigning values to directed arcs from places to transitions.

- $O = \{(t_1, p_3), (t_1, p_4), (t_2, p_3), (t_2, p_4), (t_2, p_5), (t_3, p_4), (t_3, p_5)\}$;
- $\alpha(t_1) = 0.8$, $\alpha(t_2) = 0.5$, $\alpha(t_3) = 0.2$;
- $\beta(t_1, p_3) = 0.95$, $\beta(t_1, p_4) = \beta(t_3, p_4) = 0.2$, $\beta(t_2, p_3) = \beta(t_2, p_5) = 0.1$, $\beta(t_2, p_4) = 0.92$, $\beta(t_3, p_5) = 0.94$;
- $M_0(p_1) = 0.9$, $M_0(p_2) = 1.0$, $M_0(p_3) = M_0(p_4) = M_0(p_5) = 0$;
- $w(p_1, t_1) = 0.98$, $w(p_1, t_2) = w(p_1, t_3) = w(p_2, t_1) = w(p_2, t_2) = 0.9$, $w(p_2, t_3) = 0.97$.

By incorporating $w$ with $\beta$ as stated in the above transformation, we thus get an FPN $\mathcal{N} = (P, T, I, O, \alpha, \beta_w, M_0)$, where

- $\beta_w(t_1, p_3) = \beta_w(t_2, p_4) = \beta_w(t_3, p_5) = 0.9$, $\beta(t_1, p_4) = \beta(t_3, p_4) = 0.2$, $\beta(t_2, p_3) = \beta(t_2, p_5) = 0.1$.

This FPN is none other than the one in Example 1; it is clear that $\mathcal{N}$ has the same state transition mechanism as $\mathcal{N}_w$.

Since we are looking at FPNs as a modeling formalism for CW, as was the focus for fuzzy automata in [6], we need to specify precisely the language accepted by an FPN. Suppose that $\widetilde{\Sigma}$ is a finite set of words and whose (fuzzy) language is to be modeled by an FPN. Then it is necessary to specify what word each transition corresponds to, and moreover, in order to define the language accepted by an FPN, we need a notion of "final states" which is analogous to the notion of final states in automata theory. To this end, we formalize fuzzy Petri nets for CW as follows.

*Definition 5:* A *fuzzy Petri net for computing with words* (or FPNCW for short) is defined as a ten-tuple

$$\widetilde{\mathcal{N}} = (P, T, I, O, \alpha, \beta, M_0, M_1, \widetilde{\Sigma}, l),$$

where

1) $(P, T, I, O, \alpha, \beta, M_0)$ is an FPN.
2) $M_1 : P \longrightarrow [0, 1]$ is a *final marking*, representing the final state of the net. For each $p_i \in P$, $M_1(p_i)$ indicates the degree to which $p_i$ is a final state.
3) $\widetilde{\Sigma}$ is a finite set of words for transition labeling, namely, $\widetilde{\Sigma} \subseteq \mathcal{F}(\Sigma)$, where $\Sigma$ is a finite set of symbols.
4) $l : T \longrightarrow \widetilde{\Sigma}$ is a transition labeling function.

In the graphs of FPNCWs, we put labels over the corresponding transitions; for the final marking, whenever $M_1(p_i) > 0$, we mark the place $p_i$ by a double circle and write the value $M_1(p_i)$ around the double circle.

$$f_{\mathcal{N}_w}(M, t_j)(p_k) = \begin{cases} M(p_k) \vee \left[ \mu_{M,t_j} \wedge \left( \bigwedge_{p_s \in I(t_j)} w(p_s, t_j) \right) \wedge \beta(t_j, p_k) \right], & \text{if } p_k \notin I(t_j) \\ \mu_{M,t_j} \wedge \left( \bigwedge_{p_s \in I(t_j)} w(p_s, t_j) \right) \wedge \beta(t_j, p_k), & \text{if } p_k \in I(t_j) \cap O(t_j) \\ 0, & \text{if } p_k \in I(t_j) \setminus O(t_j) \end{cases}$$

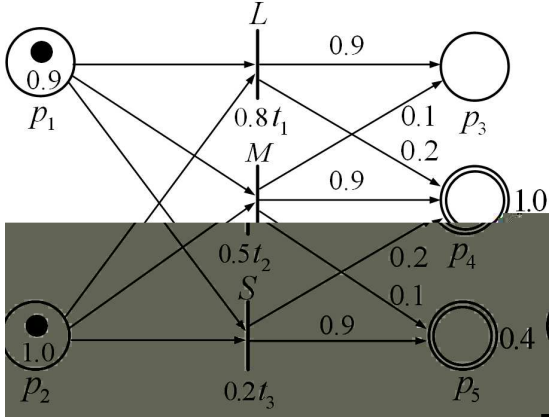

Fig. 4. An FPNCW arising from the FPN in Example 1.

The *language* $L_{\widetilde{\mathcal{N}}}$ accepted by $\widetilde{\mathcal{N}}$ is a fuzzy subset of $\widetilde{\Sigma}^*$ with the membership function defined by

$$L_{\widetilde{\mathcal{N}}}(S) = \text{height}\left[\left( \bigcup_{t \in T_S^*} f(M_0, t) \right) \cap M_1 \right]$$

for all $S \in \widetilde{\Sigma}^*$, where $T_S^*$ stands for $\{t \in T^* : l(t) = S\}$. (This uses the extended form of $l : T^* \longrightarrow \widetilde{\Sigma}^*$, which is done in the usual manner.) The membership $L_{\widetilde{\mathcal{N}}}(S)$ is the degree to which $S$ is accepted by $\widetilde{\mathcal{N}}$.

The following is an FPNCW arising from the FPN in Example 1.

*Example 4:* Let us revisit the FPN in Example 1. Suppose that we are associating the flux of hot water with each transition, where the flux of hot water is described by linguistic expressions (namely, words): $L = large$, $M = medium$, and $S = small$. More explicitly, these words interpreted as fuzzy sets are defined as follows:

$$L = large = \frac{0.1}{3} + \frac{0.6}{4} + \frac{1}{5},$$
$$M = medium = \frac{0.2}{2} + \frac{1}{3} + \frac{0.2}{4},$$
$$S = small = \frac{1}{1} + \frac{0.6}{2} + \frac{0.1}{3},$$

where the underlying input alphabet $\Sigma$ consists of discretized flux, i.e., $\Sigma = \{1, 2, 3, 4, 5\}$. Let $\widetilde{\Sigma} = \{L, M, S\}$ and define the transition labeling function $l$ by

$$l(t_1) = L, \; l(t_2) = M, \; l(t_3) = S.$$

Taking $M_1 = [0, 0, 0, 1.0, 0.4]$ as a final marking, we thus get an FPNCW $\widetilde{\mathcal{N}} = (P, T, I, O, \alpha, \beta, M_0, M_1, \widetilde{\Sigma}, l)$ (see Fig. 4), where $(P, T, I, O, \alpha, \beta, M_0)$ is the FPN in Example 1.

According to the state transition mechanism, we can compute the language accepted by $\widetilde{\mathcal{N}}$. For example, we have that

$$\begin{aligned} L_{\widetilde{\mathcal{N}}}(M) &= \text{height}\left[\left( \bigcup_{t \in T_M^*} f(M_0, t) \right) \cap M_1 \right] \\ &= \text{height}[f(M_0, t_2) \cap M_1] \\ &= \text{height}\big([0, 0, 0.1, 0.9, 0.1] \cap [0, 0, 0, 1.0, 0.4]\big) \\ &= 0.9 \end{aligned}$$

and

$$\begin{aligned} L_{\widetilde{\mathcal{N}}}(MS) &= \text{height}\left[\left( \bigcup_{t \in T_{MS}^*} f(M_0, t) \right) \cap M_1 \right] \\ &= \text{height}[f(M_0, t_2 t_3) \cap M_1] \\ &= \text{height}\big([0, 0, 0, 0, 0] \cap [0, 0, 0, 1.0, 0.4]\big) \\ &= 0. \end{aligned}$$

In other words, the degrees to which the strings "$M$" and "$MS$" are accepted are 0.9 and 0, respectively.

We end this section with a remark.

*Remark 3:* Inheriting Petri nets, FPNCWs are oriented towards capturing the concurrent nature of separate sub-models forming a complete model. The combination of two such asynchronous sub-models based on automaton models tends to become complex and hide some of the intuitive structure involved in this combination, while FPNCWs form a much more natural framework for these situations, and make it easier to visualize this structure. For FPNCWs, the combination involves simply duplicating the input tokens with their labels, feeding these into each component FPNCW, and simply selecting the appropriate output places and the associated fuzzy truth values. Due to the limit of space, we do not illustrate these claims by examples here; the reader may use the combination of original Petri nets [31] for reference.

## III. COMPUTING WITH MORE WORDS

In the last section, we have proposed a Petri net model of CW. The most distinctive feature of this model is that the label set $\widetilde{\Sigma}$ of an FPNCW consists of some words (i.e., fuzzy subsets of $\Sigma$); for instance, $\widetilde{\Sigma} = \{L, M, S\}$ in Example 4. In the same example, we have seen that $L_{\widetilde{\mathcal{N}}}(M) = 0.9$. However, if we consider the word $M' = almost\ Medium$ which is very similar to $M = Medium$, then it is easy to see that $L_{\widetilde{\mathcal{N}}}(M') = 0$ just since $M' \notin \widetilde{\Sigma}$. It seems not reasonable to discriminate strictly between similar words, because the fuzzy sets in $\widetilde{\Sigma}$ are mathematical expressions of linguistic terms that are usually selected by an expert and are always somewhat imprecise and vague. Thus, if the word $W$ is associated to a transition that can fire, we hope that $W'$ would be associated to some transition that can also fire whenever $W'$ is similar to $W$. In other words, we wish to extend $\widetilde{\Sigma}$. To this end, we have to extend related components in the FPNCW. We approach this by exploiting fuzzy reasoning.

This section consists of three subsections which are devoted to building a rule base, developing a reasoning algorithm, and investigating the formal languages represented by the extended model, respectively.

*A. Building Rule Base*

To do so, let us first recall the general scheme of fuzzy reasoning. Fuzzy reasoning is based upon a set of fuzzy IF-THEN rules, called *rule base*, also referred to as *knowledge base*. A fuzzy IF-THEN rule expresses a fuzzy implication relation between the fuzzy sets of the antecedents and the fuzzy sets of the consequents. There are a large number of models for fuzzy reasoning (see, for example, [37]). For simplicity, in the paper we consider only the Mamdani's fuzzy model [21]. In this model, the rule base is constituted by some fuzzy IF-THEN rules of the following form:

IF $x_1$ is $A_1$ AND $x_2$ is $A_2$ AND $\cdots$ AND $x_q$ is $A_q$, THEN $y$ is $B$.

The linguistic variables $x_i$, $i = 1, 2, \ldots, q$, are the antecedents and the linguistic variable $y$ is the consequent. AND is a connective operator between fuzzy concepts and it is generally implemented by means of a $t$-norm, usually the min operator $\wedge$. For example, a fuzzy IF-THEN rule for a heating management could be

IF *temperature* is *Cold* AND *oil* is *Cheap*, THEN *heating* is *High*

where *Cold*, *Cheap*, and *High* are fuzzy sets defined on corresponding universal sets.

The various rules of a set of fuzzy IF-THEN rules are generally joined by an $s$-norm, generally the union operator $\vee$, and all together form the fuzzy algorithm. Note that in a fuzzy IF-THEN rule, none of the antecedents has independent relation with the consequent: It is only the intersection of the antecedents that has the relation.

In order to draw conclusions from a rule base, we need an inference mechanism that can produce an output from a collection of fuzzy IF-THEN rules. The most famous rule of inference is the generalized modus ponens. In the general case of $p$ fuzzy IF-THEN rules with $q$ antecedents, the fuzzy inference engine reads like:

- Rule 1: IF $x_1$ is $A_{11}$ AND $x_2$ is $A_{12}$ AND $\cdots$ AND $x_q$ is $A_{1q}$, THEN $y$ is $B_1$.
  $\vdots$
- Rule $i$: IF $x_1$ is $A_{i1}$ AND $x_2$ is $A_{i2}$ AND $\cdots$ AND $x_q$ is $A_{iq}$, THEN $y$ is $B_i$.
  $\vdots$
- Rule $p$: IF $x_1$ is $A_{p1}$ AND $x_2$ is $A_{p2}$ AND $\cdots$ AND $x_q$ is $A_{pq}$, THEN $y$ is $B_p$.
- Fact: $x_1$ is $A'_1$ AND $x_2$ is $A'_2$ AND $\cdots$ AND $x_q$ is $A'_q$.
- Conclusion: $y$ is $B'$.

Using Mamdani's max-min fuzzy implication rule, the generalized modus ponens inference procedure gives rise to $B'(y) = \vee_{i=1}^{p} B'_i(y)$, where $B'_i(y) = \wedge_{j=1}^{q} \{\vee_{x_j} [A_{ij}(x_j) \wedge A'_j(x_j) \wedge B_i(y)]\}$.

Let $\widetilde{\mathcal{N}} = (P, T, I, O, \alpha, \beta, M_0, M_1, \widetilde{\Sigma}, l)$ be an FPNCW. Since our aim is to use $\widetilde{\mathcal{N}}$ for computing with more words, we are going to extend $\widetilde{\Sigma}$ to $\widetilde{\Sigma}'$. To this end, we need to reason the system behavior when firing the transitions labeled with the new words in $\widetilde{\Sigma}'$ and extend the related components of $\widetilde{\mathcal{N}}$. Let us begin with the rule base arising from $\widetilde{\mathcal{N}}$. Assume that the transition $t_j$ labeled $l(t_j)$ has input places $p_{j1}, \ldots, p_{j|I(t_j)|}$ (where the notation $|I(t_j)|$ denotes the cardinality of $I(t_j)$) and each of these places has a token labeled 1.0. Then $t_j$ can fire and gives a state distribution $\sum_{i=1}^{m} \beta(t_j, p_i)/p_i$ on all places. According to this, we associate to each transition $t_j$ a fuzzy IF-THEN rule $\mathrm{R}_{t_j}$:

IF $p_{j1}$ is $1/p_{j1}$ AND $\cdots$ AND $p_{j|I(t_j)|}$ is $1/p_{j|I(t_j)|}$ AND *label of transition* is $l(t_j)$, THEN *next state distribution* is $D_{t_j} = \sum_{i=1}^{m} \beta(t_j, p_i)/p_i$.

Here, $1/p_{js}$, $1 \leq s \leq |I(t_j)|$, is a singleton in $P$, i.e., the fuzzy subset of $P$ with membership 1 at $p_{js}$ and with zero membership for all the other elements of $P$. The rule base associated to $\widetilde{\mathcal{N}}$, denoted by $\mathfrak{R}$, consists of all such fuzzy IF-THEN rules $\mathrm{R}_{t_j}$. Clearly, there are only $|T|$ rules in $\mathfrak{R}$.

For subsequent need, let us build the rule base associated to the FPNCW in Example 4.

*Example 5:* Let $\widetilde{\mathcal{N}}$ be the FPNCW in Example 4. Following the above method of building rule bases associated to FPNCWs, we see that $\mathfrak{R}$ of $\widetilde{\mathcal{N}}$ consists of three fuzzy IF-THEN rules:

$\mathrm{R}_{t_1}$: IF $p_1$ is $1/p_1$ AND $p_2$ is $1/p_2$ AND *label of transition* is $L$, THEN *next state distribution* is $D_{t_1} = 0.9/p_3 + 0.2/p_4$.

$\mathrm{R}_{t_2}$: IF $p_1$ is $1/p_1$ AND $p_2$ is $1/p_2$ AND *label of transition* is $M$, THEN *next state distribution* is $D_{t_2} = 0.1/p_3 + 0.9/p_4 + 0.1/p_5$.

$\mathrm{R}_{t_3}$: IF $p_1$ is $1/p_1$ AND $p_2$ is $1/p_2$ AND *label of transition* is $S$, THEN *next state distribution* is $D_{t_3} = 0.2/p_4 + 0.9/p_5$.

For instance, consider the fact that $p_1$ is $1/p_1$ AND $p_2$ is $1/p_2$ AND *label of transition* is $L'$, where $L'$ is *almost large* defined by the membership function $L'(x) = [L(x)]^{\frac{1}{2}}$ for any $x \in \Sigma = \{1, 2, 3, 4, 5\}$. By a simple calculation, we see that $L' = 0.32/3 + 0.77/4 + 1/5$. We obtain by fuzzy reasoning that *next state distribution* $D_{L'}$ is

$$\begin{aligned}
D_{L'}(p_1) &= D'_{t_1}(p_1) \vee D'_{t_2}(p_1) \vee D'_{t_3}(p_1) \\
&= 0 \vee 0 \vee 0 = 0; \\
D_{L'}(p_2) &= D'_{t_1}(p_2) \vee D'_{t_2}(p_2) \vee D'_{t_3}(p_2) \\
&= 0 \vee 0 \vee 0 = 0; \\
D_{L'}(p_3) &= D'_{t_1}(p_3) \vee D'_{t_2}(p_3) \vee D'_{t_3}(p_3) \\
&= 0.9 \vee 0.1 \vee 0 = 0.9; \\
D_{L'}(p_4) &= D'_{t_1}(p_4) \vee D'_{t_2}(p_4) \vee D'_{t_3}(p_4) \\
&= 0.2 \vee 0.32 \vee 0.1 = 0.32; \\
D_{L'}(p_5) &= D'_{t_1}(p_5) \vee D'_{t_2}(p_5) \vee D'_{t_3}(p_5) \\
&= 0 \vee 0.1 \vee 0.1 = 0.1,
\end{aligned}$$

namely, $D_{L'} = 0.9/p_3 + 0.32/p_4 + 0.1/p_5$.

Similarly, define $M' = almost\ medium = 0.45/2 + 1/3 + 0.45/4$ and $S' = almost\ small = 1/1 + 0.77/2 + 0.32/3$. Then given the fact that $p_1$ is $1/p_1$ AND $p_2$ is $1/p_2$ AND *label of transition* is $M'$ (resp. $S'$), we can get by a routine computation that $D_{M'} = 0.45/p_3 + 0.9/p_4 + 0.45/p_5$ (resp. $D_{S'} = 0.1/p_3 + 0.32/p_4 + 0.9/p_5$).

### B. Algorithm for Extending FPNCWs

Having built the rule base, we can now turn to the extension of FPNCWs. The basic idea behind the extension is to keep places and increase transitions by fuzzy reasoning. The resultant FPNs are referred to as *fuzzy Petri nets for computing with more words* (FPNCMWs).

For later need, let us partition the $|T|$ rules in $\mathfrak{R}$ into (mutually exclusive) groups such that all rules in a group have the same antecedents. Suppose that there are $k$ groups consisting of $\mathfrak{R}_1, \ldots, \mathfrak{R}_k$ and the transitions related to the rules in a group $\mathfrak{R}_i$ are $t_{i1}, \ldots, t_{in_i}$. Hence, $\sum_{i=1}^{k} n_i = |T|$ and $I(t_{i1}) = \cdots = I(t_{in_i})$. In other words, the transitions $t_{i1}, \ldots, t_{in_i}$ have the common input places, say, $p_{i1}, \ldots, p_{im_i}$, and we write $I(\mathfrak{R}_i)$ for the set of these places. In addition, for any $\mathfrak{R}' \subseteq \mathfrak{R}$ we write $T(\mathfrak{R}')$ for the set of transitions related to the rules in $\mathfrak{R}'$.

*Algorithm for Extending FPNCWs:*

INPUT: FPNCW $\widetilde{\mathcal{N}} = (P, T, I, O, \alpha, \beta, M_0, M_1, \widetilde{\Sigma}, l)$ and $\widetilde{\Sigma}'$, where $|\widetilde{\Sigma}'| < \infty$ and $\widetilde{\Sigma} \subseteq \widetilde{\Sigma}' \subseteq \mathcal{F}(\Sigma)$.

OUTPUT: FPNCMW $\widetilde{\mathcal{N}}' = (P', T', I', O', \alpha', \beta', M_0', M_1', \widetilde{\Sigma}', l')$.

PROCEDURE:

Step 1) Let $P' = P, M_0' = M_0$, and $M_1' = M_1$; build and partition the rule base $\mathfrak{R}$ associated to $\widetilde{\mathcal{N}}$.

Step 2) For each $\mathfrak{R}_i$ ($1 \leq i \leq k$) and each $W_j' \in \widetilde{\Sigma}' \setminus \widetilde{\Sigma} = \{W_1', \ldots, W_r'\}$, define
$$\mathfrak{R}_{ij} := \{R_{t_s} \in \mathfrak{R}_i : height(l(t_s) \cap W_j') > 0\}.$$

Step 3) For each $1 \leq i \leq k$ and $1 \leq j \leq r$, if $\mathfrak{R}_{ij} \neq \emptyset$, then it is associated with a transition $t_{ij}'$. Set $T' = T \bigcup \{t_{ij}' : \mathfrak{R}_{ij} \neq \emptyset\}$.

Step 4) For each $t_{ij}' \in T'$, take $I'(t_{ij}') = I(\mathfrak{R}_i)$; for each $t_j \in T$, take $I'(t_j) = I(t_j)$. Define $(p', t') \in I'$ if and only if $p' \in I'(t')$.

Step 5) For each $t_{ij}' \in T'$, take $O'(t_{ij}') = \{p \in P : \exists R_{t_s} \in \mathfrak{R}_{ij}$ such that $D_{t_s}(p) > 0\}$; for each $t_j \in T$, take $O'(t_j) = O(t_j)$. Define $(t', p') \in O'$ if and only if $p' \in O'(t')$.

Step 6) For each $t_{ij}' \in T'$, take $\alpha'(t_{ij}') = \vee\{\alpha(t_s) : t_s \in T(\mathfrak{R}_{ij})\}$; for each $t_j \in T$, take $\alpha'(t_j) = \alpha(t_j)$.

Step 7) For each $t_{ij}' \in T'$, we derive a state distribution $D_{t_{ij}'}$ by applying fuzzy reasoning to the rules in $\mathfrak{R}_{ij}$ and the following fact:

$p_{i1}$ is $1/p_{i1}$ AND $\cdots$ AND $p_{im_i}$ is $1/p_{im_i}$ AND *label of transition* is $W_j'$.

For any $(t', p') \in O'$, if $t' = t_{ij}'$ for some $i$ and $j$, then set $\beta'(t', p') = D_{t_{ij}'}(p')$; otherwise, set $\beta'(t', p') = \beta(t', p')$.

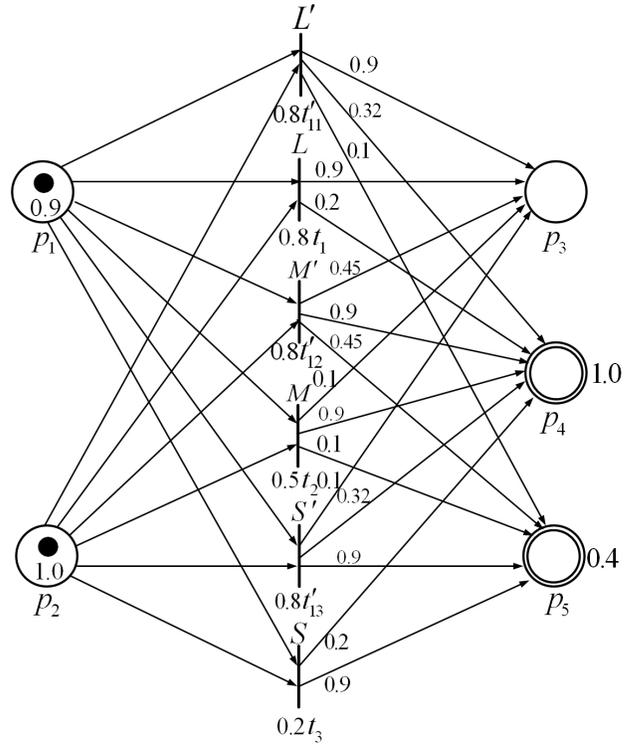

Fig. 5. An FPNCMW $\widetilde{\mathcal{N}}'$ arising from the FPNCW $\widetilde{\mathcal{N}}$ in Example 4.

Step 8) For each $t_{ij}' \in T'$, define $l'(t_{ij}') = W_j'$; for each $t_j \in T$, define $l'(t_j) = l(t_j)$.

Let us give a brief, informal account of the above steps. Since we are concerned with the extension of transition labels, we keep the places, and thus keep the initial and final markings. Observe that if two rules in the rule base $\mathfrak{R}$ associated to $\widetilde{\mathcal{N}}$ have different antecedents on places, then they necessarily yield a trivial inference conclusion. This forces us partition $\mathfrak{R}$ according to whether or not two rules involve the same places. In Step 2), we are paying attention to the fuzzy IF-THEN rules whose labels of transitions match $W_j'$ with nonzero degree since, as we have seen from the previous inference procedure, only these rules contribute to our inference conclusion. The choice of threshold values in Step 6) implies that whenever $t_{ij}'$ can fire, all transitions related to it can fire as well. This choice makes our inference somewhat simple; in fact, choosing smaller threshold values are also feasible. The remaining steps depend upon our fuzzy reasoning mechanism; they are natural and comprehensible.

It follows immediately from the above algorithm that $\widetilde{\mathcal{N}}'$ has $\widetilde{\mathcal{N}}$ as a full subnet. More explicitly, $P = P', T \subseteq T', I \subseteq I', O \subseteq O', \alpha = \alpha'|_T, \beta = \beta'|_O, M_0 = M_0', M_1 = M_1', \widetilde{\Sigma} \subseteq \widetilde{\Sigma}', l = l'|_T$, where the notation $\varphi|_{X'}$ means that we are restricting the mapping $\varphi$ defined on $X$ to the smaller domain $X'$.

Let us use an example to illustrate the algorithm.

*Example 6:* Let $\widetilde{\mathcal{N}}$ be the FPNCW in Example 4 and $\widetilde{\Sigma}' = \widetilde{\Sigma} \bigcup \{L', M', S'\}$, where $L', M'$, and $S'$ are defined in Example 5. With the results of Examples 4 and 5, we can derive an FPNCMW $\widetilde{\mathcal{N}}' = (P', T', I', O', \alpha', \beta', M_0', M_1', \widetilde{\Sigma}', l')$ as follows.

$$f'(M,t')(p) = \begin{cases} M(p) \vee \bigvee_{t_s \in T(\mathfrak{R}_{ij})} \left[\text{height}\left(l(t_s) \cap l(t'_{ij})\right) \wedge f(M,t_s)(p)\right], & \text{if } p \notin I'(t') \\ \bigvee_{t_s \in T(\mathfrak{R}_{ij})} \left[\text{height}\left(l(t_s) \cap l(t'_{ij})\right) \wedge f(M,t_s)(p)\right], & \text{if } p \in I'(t') \end{cases}$$

Step 1) Let $P' = \{p_1, p_2, p_3, p_4, p_5\}$, $M'_0 = [0.9, 1.0, 0, 0, 0]$, and $M'_1 = [0, 0, 0, 1.0, 0.4]$. The rule base $\mathfrak{R}$ associated to $\widetilde{\mathcal{N}}$ has been built in Example 5. Furthermore, it is easy to see that all the three rules belong to the same group, that is, $k = 1$.

Step 2) For $\mathfrak{R}_1 (= \mathfrak{R})$ and $L'$ (resp. $M'$, $S'$), we have that $\mathfrak{R}_{11} = \mathfrak{R}$ (resp. $\mathfrak{R}_{12} = \mathfrak{R}_{13} = \mathfrak{R}$).

Step 3) Set $T' = \{t_1, t_2, t_3, t'_{11}, t'_{12}, t'_{13}\}$.

Step 4) For any $t' \in T'$, take $I'(t') = \{p_1, p_2\}$ and $I' = I \bigcup \{(p_1, t'_{11}), (p_2, t'_{11}), (p_1, t'_{12}), (p_2, t'_{12}), (p_1, t'_{13}), (p_2, t'_{13})\}$.

Step 5) For any $t' \in T'$, take $O'(t') = \{p_3, p_4, p_5\}$ and $O' = O \bigcup \{(t'_{11}, p_3), (t'_{11}, p_4), (t'_{11}, p_5), (t'_{12}, p_3), (t'_{12}, p_4), (t'_{12}, p_5), (t'_{13}, p_3), (t'_{13}, p_4), (t'_{13}, p_5)\}$.

Step 6) Take $\alpha'(t_1) = \alpha(t_1) = 0.8$, $\alpha'(t_2) = \alpha(t_2) = 0.5$, $\alpha'(t_3) = \alpha(t_3) = 0.2$, and $\alpha'(t'_{11}) = \alpha'(t'_{12}) = \alpha'(t'_{13}) = 0.8$.

Step 7) It follows from Example 5 that

$$D_{t'_{11}} = D_{L'} = 0.9/p_3 + 0.32/p_4 + 0.1/p_5;$$
$$D_{t'_{12}} = D_{M'} = 0.45/p_3 + 0.9/p_4 + 0.45/p_5;$$
$$D_{t'_{13}} = D_{S'} = 0.1/p_3 + 0.32/p_4 + 0.9/p_5.$$

We thus set $\beta'(t_1, p_3) = \beta'(t_2, p_4) = \beta'(t_3, p_5) = \beta'(t'_{11}, p_3) = \beta'(t'_{12}, p_4) = \beta'(t'_{13}, p_5) = 0.9$, $\beta'(t'_{12}, p_3) = \beta'(t'_{12}, p_5) = 0.45$, $\beta'(t'_{11}, p_4) = \beta'(t'_{13}, p_4) = 0.32$, $\beta'(t_1, p_4) = \beta'(t_3, p_4) = 0.2$, and $\beta'(t_2, p_3) = \beta'(t_2, p_5) = \beta'(t'_{11}, p_5) = \beta'(t'_{13}, p_3) = 0.1$.

Step 8) Define $l'(t_1) = L$, $l'(t_2) = M$, $l'(t_3) = S$, $l'(t'_{11}) = L'$, $l'(t'_{12}) = M'$, and $l'(t'_{13}) = S'$.

Finally, the FPNCMW $\widetilde{\mathcal{N}}'$ is depicted in Fig. 5.

### C. Languages of FPNCMWs

Let $\widetilde{\mathcal{N}} = (P, T, I, O, \alpha, \beta, M_0, M_1, \widetilde{\Sigma}, l)$ be an FPNCW and $\widetilde{\mathcal{N}}' = (P, T', I', O', \alpha', \beta', M_0, M_1, \widetilde{\Sigma}', l')$ be an extension of $\widetilde{\mathcal{N}}$ for computing with more words. Suppose that $f : [0,1]^n \times T \longrightarrow [0,1]^n$ and $f' : [0,1]^n \times T' \longrightarrow [0,1]^n$ are state transition functions of $\widetilde{\mathcal{N}}$ and $\widetilde{\mathcal{N}}'$, respectively. We are ready to clarify some relationships between $f'$ and $f$.

Let us first consider when $f'$ is defined for a transition $t' \in T'$. Assume that the current marking of $\widetilde{\mathcal{N}}'$ is $M$. If $t' = t_j \in T$ for some $j$, then it follows from definition that $f'$ is defined for $t'$ if and only if $f$ is defined for $t'$, because all data associated to $t_j$ have not been modified in the extending. Otherwise, we have that $t' = t'_{ij}$ and there is a set $\mathfrak{R}_{ij}$ of rules for some $i$ and $j$. By definition, $f'$ is defined for $t'_{ij}$ if and only if $M(p) \geq \alpha(t'_{ij}) = \vee\{\alpha(t_s) : t_s \in T(\mathfrak{R}_{ij})\}$ for every $p \in I'(t'_{ij})$. This means that $f'$ is defined for $t'_{ij}$ if and only if for any $t_s \in T(\mathfrak{R}_{ij})$, $M(p) \geq \alpha(t_s)$ for every $p \in I(t_s)$, since $I'(t'_{ij}) = I(t_s)$. In other words, $f'$ is defined for $t'_{ij}$ if and only if $f$ is defined for all $t_s \in T(\mathfrak{R}_{ij})$.

Further, if $f'(M, t')$ is defined, then we have the observation below.

*Theorem 1:* For any $t' \in T'$, we have the following:
1) If $t' = t_j \in T(\subseteq T')$ for some $j$, then $f'(M, t') = f(M, t_j)$.
2) If $t' = t'_{ij} \in T' \setminus T$ for some $i$ and $j$, then we have the equation at the top of this page.

*Proof:* See Appendix A. ∎

As an immediate consequence of the above theorem, we have the following.

*Corollary 1:*
1) The FPNCMW $\widetilde{\mathcal{N}}'$ is a faithful extension of $\widetilde{\mathcal{N}}$ in the sense that $f'(M, t) = f(M, t)$ for any $t \in T$.
2) For any $S \in \widetilde{\Sigma}^*$, $L_{\widetilde{\mathcal{N}}'}(S) = L_{\widetilde{\mathcal{N}}}(S)$.

## IV. LANGUAGE EXPRESSIVENESS OF THE TWO FORMAL MODELS OF COMPUTING WITH WORDS

In the previous sections, we have established a formal model of CW based on FPNs. Recall that based on fuzzy automata, another formal model of CW was proposed and investigated in [44] and [6]. To compare the language expressiveness of the two formal models, let us review two definitions from [6].

*Definition 6:* A *fuzzy automaton for computing with words* (or FACW for short) is a five-tuple $M = (Q, \widetilde{\Sigma}, \delta, q_0, F)$, where:
1) $Q$ is a finite set of states.
2) $\widetilde{\Sigma}$ is a subset of $\mathcal{F}(\Sigma)$, where $\Sigma$ is a finite set of symbols, called the underlying input alphabet.
3) $q_0$, a member of $Q$, is the initial state.
4) $F$ is a fuzzy subset of $Q$, called the fuzzy set of final states and for each $q \in Q$, $F(q)$ indicates intuitively the degree to which $q$ is a final state.
5) $\delta$ is a fuzzy transition function from $Q \times \widetilde{\Sigma}$ to $\mathcal{F}(Q)$ that takes a state in $Q$ and a word in $\widetilde{\Sigma}$ as arguments and returns a fuzzy subset of $Q$.

For any $p, q \in Q$ and $W \in \widetilde{\Sigma}$, we may interpret $\delta(p, W)(q)$ as the possibility degree to which the automaton in state $p$ and with input $W$ may enter state $q$.

To describe what happens when inputting a sequence of words, let us extend the fuzzy transition function to strings.

*Definition 7:* Let $M = (Q, \widetilde{\Sigma}, \delta, q_0, F)$ be an FACW.
1) The *extended fuzzy transition function* from $Q \times \widetilde{\Sigma}^*$ to $\mathcal{F}(Q)$, denoted by the same notation $\delta$, is defined inductively as follows:

$$\delta(p, \epsilon) = 1/p$$
$$\delta(p, SW) = \cup_{q \in Q}[\delta(p, S)(q) \cdot \delta(q, W)]$$

for all $S \in \widetilde{\Sigma}^*$ and $W \in \widetilde{\Sigma}$, where $\delta(p, S)(q) \cdot \delta(q, W)$ stands for the scale product of the membership $\delta(p, S)(q)$ and the fuzzy set $\delta(q, W)$.

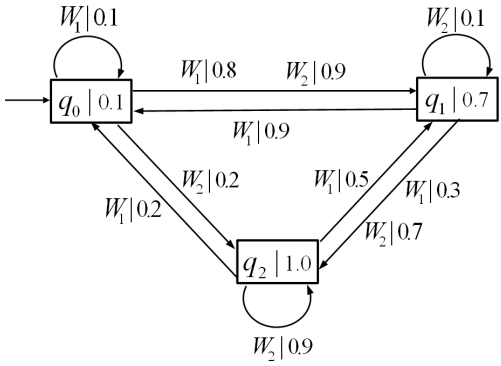

Fig. 6. An example of FACWs.

2) The *language* $L_M$ accepted by $M$ is a fuzzy subset of $\widetilde{\Sigma}^*$ with the membership function defined by

$$L_M(S) = \text{height}(\delta(q_0, S) \cap F)$$

for all $S \in \widetilde{\Sigma}^*$. The membership $L_M(S)$ is the degree to which $S$ is accepted by $M$.

For subsequent need, we record a very simple example of FACWs.

*Example 7:* Let $Q = \{q_0, q_1, q_2\}$ and $\widetilde{\Sigma} = \{W_1, W_2\}$, where

$$W_1 = \frac{1}{1} + \frac{0.5}{2} + \frac{0.1}{3},$$
$$W_2 = \frac{0.1}{3} + \frac{0.5}{4} + \frac{1}{5}$$

with the underlying input alphabet $\Sigma = \{1, 2, 3, 4, 5\}$. Let $q_0$ be the initial state and $F = 0.1/q_0 + 0.7/q_1 + 1.0/q_2$ be the fuzzy set of final states. The fuzzy transition function $\delta$ is depicted in Fig. 6, where an arc from $q_i$ to $q_j$ with label $W|x$ means that $\delta(q_i, W)(q_j) = x$. A box with $q_i|y$ in Fig. 6 means that $F(q_i) = y$. We thus get an FACW $M = (Q, \widetilde{\Sigma}, \delta, q_0, F)$. According to the fuzzy transition function, we can compute the language accepted by $M$. For example, the degree to which the string "$W_1 W_2$" is accepted is 0.7.

We can now consider the class of fuzzy languages that can be represented by each formalism. Recall that in classical models of computation, the class of Petri net languages is strictly larger than the class of regular languages, meaning that Petri nets with finite sets of places and transitions can represent more languages than finite-state automata [31]. With a little surprise, we claim that FACWs and FPNCWs are equivalent, that is, they represent the same fuzzy languages. In order to prove this result, it is sufficient to see how any FACW can always be transformed into an FPNCW that accepts the same language, and vice versa.

Suppose that we are given an FACW $M = (Q, \widetilde{\Sigma}, \delta, q_0, F)$, where $\widetilde{\Sigma}$ is a finite set. To construct an FPNCW $\widetilde{\mathcal{N}}_M = (P, T, I, O, \alpha, \beta, M_0, M_1, \widetilde{\Sigma}, l)$ such that $L_{\widetilde{\mathcal{N}}_M} = L_M$, we can proceed as follows.

(1) We first view each state in $Q$ as defining a unique place in $P$, that is, $P = Q$. This also immediately specifies:
   – the initial marking $M_0$ of $\widetilde{\mathcal{N}}_M$: for any $q \in P$, $M_0(q) = 1$ if $q = q_0$, and $M_0(q) = 0$ otherwise;
   – the finial marking $M_1 = F$.

(2) Next, we associate each triple $(q, W, q')$ satisfying $\delta(q, W)(q') > 0$ in $M$ with a transition $t_{(q,W,q')}$ in $T$ of $\widetilde{\mathcal{N}}_M$. Formally, $T = \{t_{(q,W,q')} : \exists\, q, q' \in Q, W \in \widetilde{\Sigma}$ such that $\delta(q, W)(q') > 0\}$. We then:
   – set $I = \{(q, t_{(q,W,q')}) : t_{(q,W,q')} \in T\}$;
   – set $O = \{(t_{(q,W,q')}, q') : t_{(q,W,q')} \in T\}$;
   – for any $t \in T$, take $\alpha(t) = \wedge\{\delta(q, W)(q') > 0 : q, q' \in Q, W \in \widetilde{\Sigma}\}$;
   – for any $(t_{(q,W,q')}, q') \in O$, define $\beta(t_{(q,W,q')}, q') = \delta(q, W)(q')$;
   – define $l(t_{(q,W,q')}) = W$.

By the above construction, we see that for any transition $t_j$ of $\widetilde{\mathcal{N}}_M$, there is exactly one input place and one output place associated to $t_j$. Let us give a simple example that illustrates the above construction.

*Example 8:* Consider the FACW $M$ in Example 7. We now follow the above procedure to construct an equivalent FPNCW $\widetilde{\mathcal{N}}_M = (P, T, I, O, \alpha, \beta, M_0, M_1, \widetilde{\Sigma}, l)$:

(1) Let $P = Q = \{q_0, q_1, q_2\}$, and then take $M_0 = [1, 0, 0]$ and $M_1 = [0.1, 0.7, 1.0]$.

(2) Next, we associate each triple $(q_i, W_j, q_k)$ satisfying $\delta(q_i, W_j)(q_k) > 0$ in $M$ with a transition $t_{ijk}$ in $T$ of $\widetilde{\mathcal{N}}_M$, that is, $T = \{t_{010}, t_{011}, t_{021}, t_{022}, t_{110}, t_{121}, t_{112}, t_{122}, t_{210}, t_{211}, t_{222}\}$. We then:
   – set $I = \{(q_0, t_{010}), (q_0, t_{011}), (q_0, t_{021}), (q_0, t_{022}), (q_1, t_{110}), (q_1, t_{121}), (q_1, t_{112}), (q_1, t_{122}), (q_2, t_{210}), (q_2, t_{211}), (q_2, t_{222})\}$;
   – set $O = \{(t_{010}, q_0), (t_{011}, q_1), (t_{021}, q_1), (t_{022}, q_2), (t_{110}, q_0), (t_{121}, q_1), (t_{112}, q_2), (t_{122}, q_2), (t_{210}, q_0), (t_{211}, q_1), (t_{222}, q_2)\}$;
   – for any $t \in T$, take $\alpha(t) = 0.1$;
   – for any $(t_{ijk}, q_k) \in O$, define $\beta(t_{ijk}, q_k) = \delta(q_i, W_j)(q_k)$;
   – define $l(t_{ijk}) = W_j$.

The FPNCW $\widetilde{\mathcal{N}}_M$ is depicted in Fig. 7; it is not difficult for us to check that $L_{\widetilde{\mathcal{N}}_M} = L_M$.

More generally, we have the following result.

*Proposition 1:* Let $M$ and $\widetilde{\mathcal{N}}_M$ be as in the above construction. Then $L_{\widetilde{\mathcal{N}}_M} = L_M$.

*Proof:* See Appendix A. ∎

Conversely, we are ready to show how any FPNCW can be transformed into an FACW that accepts the same language. Let us begin with an observation that the reachable state set $R(\widetilde{\mathcal{N}})$ of an arbitrary FPNCW $\widetilde{\mathcal{N}}$ is finite. This is because (i) the transitions in $\widetilde{\mathcal{N}}$ are finite and (ii) the max and min operations cannot introduce a membership grade not already assigned to some place or directed arc.

Assume that we are given an FPNCW $\widetilde{\mathcal{N}} = (P, T, I, O, \alpha, \beta, M_0, M_1, \widetilde{\Sigma}, l)$. To construct an FACW $M_{\widetilde{\mathcal{N}}} = (Q, \widetilde{\Sigma}, \delta, q_0, F)$ such that $L_{M_{\widetilde{\mathcal{N}}}} = L_{\widetilde{\mathcal{N}}}$, we can follow the steps below:

(1) We regard each reachable state of $\widetilde{\mathcal{N}}$ as a state in $Q$, namely, $Q = R(\widetilde{\mathcal{N}})$. Further, we specify:
   – the initial state $q_0 = M_0$;
   – $F : Q \longrightarrow [0, 1]$, the fuzzy set of finial states, assigns to any $q \in Q$ the final state degree height$(q \cap M_1)$.

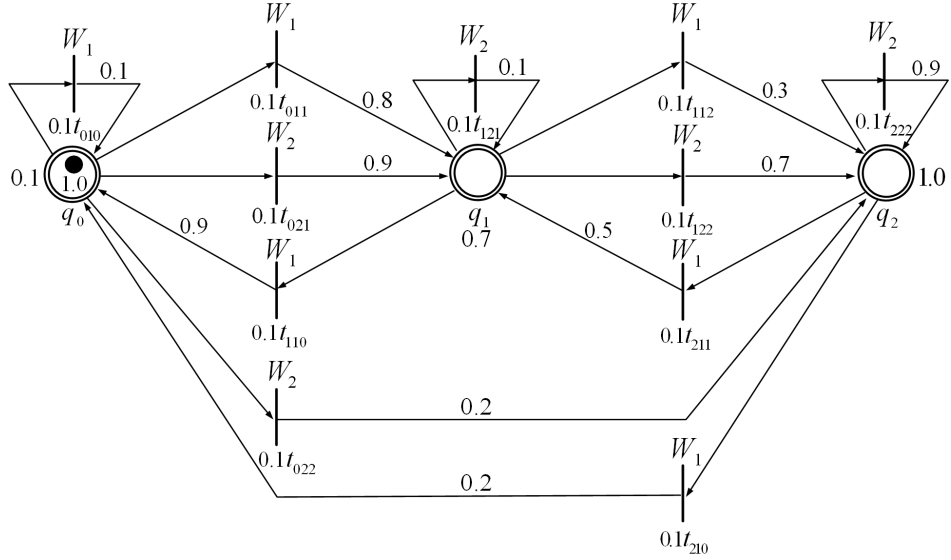

Fig. 7. An FPNCW $\widetilde{\mathcal{N}}_M$ that is equivalent to the FACW $M$ in Example 7.

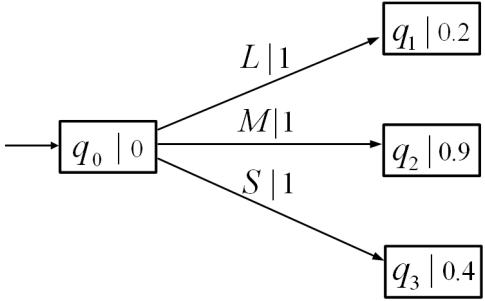

Fig. 8. An FACW $M_{\widetilde{\mathcal{N}}}$ that is equivalent to the FPNCW $\widetilde{\mathcal{N}}$ in Example 4.

(2) Next, we define the fuzzy transition function $\delta : Q \times \widetilde{\Sigma} \longrightarrow \mathcal{F}(Q)$ as follows.

$$\delta(q, W)(q') = \begin{cases} 1, & \text{if } q' \in \{f(q, t_j) : l(t_j) = W\} \\ 0, & \text{otherwise} \end{cases}$$

for any $q' \in Q$, where $f$ is the state transition function of $\widetilde{\mathcal{N}}$.

Let us see how to transform the FPNCW $\widetilde{\mathcal{N}}$ in Example 4 into an equivalent FACW $M_{\widetilde{\mathcal{N}}}$ using the above construction.

*Example 9:* Note that after Definition 4, we have given all the reachable states of the FPN in Example 1. Clearly, they are all the reachable states of the FPNCW $\widetilde{\mathcal{N}}$ in Example 4. Therefore, we have that $Q = \{q_0, q_1, q_2, q_3\}$ with $q_0 = [0.9, 1.0, 0, 0, 0]$, $q_1 = [0, 0, 0.9, 0.2, 0]$, $q_2 = [0, 0, 0.1, 0.9, 0.1]$, and $q_3 = [0, 0, 0, 0.2, 0.9]$. Using the fact $M_1 = [0, 0, 0, 1.0, 0.4]$, we obtain that $F(q_0) = 0$, $F(q_1) = 0.2$, $F(q_2) = 0.9$, and $F(q_3) = 0.4$, namely, $F = 0.2/q_1 + 0.9/q_2 + 0.4/q_3$. The fuzzy transition function $\delta$ is simple and is depicted in Fig. 8, where we are using the same notation as specified in Example 7. Taking $M_{\widetilde{\mathcal{N}}} = (Q, \widetilde{\Sigma}, \delta, q_0, F)$, it is easy to see that $L_{M_{\widetilde{\mathcal{N}}}} = L_{\widetilde{\mathcal{N}}}$.

Analogous to Proposition 1, we have the following observation.

*Proposition 2:* Let $\widetilde{\mathcal{N}}$ and $M_{\widetilde{\mathcal{N}}}$ be as in the previous construction. Then $L_{M_{\widetilde{\mathcal{N}}}} = L_{\widetilde{\mathcal{N}}}$.

*Proof:* See Appendix A. ■

## V. DISCUSSION

Along the lines of developing a computational theory for granular computing, we have introduced a formal concurrency model of CW in this paper, where words are interpreted as fuzzy sets. To make the model robust for dealing with more words, a faithful extension has been developed by exploiting fuzzy reasoning. Our model is based upon the classical computing model, Petri nets, and has words drawn from a natural language as the objects of computation; these features make our work fall into the category of CW, according to Zadeh's explanation of CW [46], [47].

The present paper, which is the first attempt at the concurrency model of CW, has been concerned mainly with the theoretical development of FPNCWs. The practical application of FPNCWs can be accomplished in at least two ways. One approach considers FPNCWs as an event-oriented modeling tool for concurrent systems with fuzzy events; this case may not need the extension from FPNCWs to FPNCMWs. We have observed that the FPNCWs are well-suited for modeling the so-called fuzzy discrete event systems [4], [5], [16], [33]. A topic of ongoing work concerns the supervisory control theory of fuzzy discrete event systems modeled by FPNCWs. The other approach is to use FPNCWs as a tool for chain reasoning with perceptive information that can be represented by fuzzy sets. This is somewhat like the application of FPNs to knowledge reasoning in [1], [9], [15], [20]; the advantage of FPNCWs is that this makes it possible to reason about fuzzy events in a dynamic way. In this case, the extension from FPNCWs to FPNCMWs is required. For example, let us consider the catalytic reactions of ethanol by heating it with an excess of concentrated sulphuric acid. It has been known that different temperatures of heat produce different resultants:

heating ethanol at about $140°C$ gives ether and water, heating it at about $170°C$ gives ethene and water, and heating it at some in-between temperatures may produce both ether and ethene with corresponding degrees. In the framework of FPNCWs, we may model reactants and resultants as places and chemical equations as transitions, where transitions are labeled by fuzzy sets representing temperatures. Using the extension from FPNCWs to FPNCMWs, we can reasoning about the possibility degrees of resultants when heating ethanol at any in-between temperature.

Because any extended FPNCMW is strongly dependent on the underlying FPNCW, another basic problem arising in the application of FPNCWs is how to choose words as the labels of the FPNCW. Intuitively, the fuzzy sets representing the words should be dense in the sense that the supports of these fuzzy sets cover their universal set. In terms of membership functions, convex and normal fuzzy sets (A *convex and normal fuzzy set* is defined by its $\alpha$-level cuts being connected and its core being nonempty.) may be a good candidate since we have established our extension by Mamdani's fuzzy implication rule and the fuzzy sets in Mamdani's model are represented in most cases by convex and normal fuzzy sets. In the practice, membership functions occurring in the rules are usually piece-wise linear. Some interpolation approaches shown in [14] may also be helpful to this issue. In addition, words mean different things to different people [23], and besides modeling by the fuzzy sets used here, they can be modeled in many other ways such as probabilistic distributions [50], type-2 fuzzy sets, and interval type-2 fuzzy sets [25], [26]. Hence, it is desirable to provide a unified formal model of CW for various modelings of words. In [3], a unified probabilistic model is proposed for handling the words modeled by probabilistic distributions and possibility distributions; a general model remains yet to be established.

It is reasonable to ask the question: Which is a better formal model of CW, an FACW or an FPNCW? There seems to be no obvious answer to such a question, as modeling is always subject to personal biases and very frequently depends on the particular application considered. From the point of view of language expressiveness, both the models are equivalent. However, from the perspective of modeling, FACWs usually describe sequential systems with fuzzy events; concurrency in synchronous systems is usually not considered. Compared to FACWs, FPNCWs have certain advantages in some compositions of sub-models and offer compact representations of concurrent systems with fuzzy events, and moreover, the modeling of concurrent systems is usually more natural using FPNCWs. In addition, the notions of causality and independence can be best studied in FPNCWs. To characterize these differences between FACWs and FPNCWs, we are currently comparing the expressiveness of the two formal models of CW by the concept of bisimulation [35], one of the most important contributions of Concurrency Theory to Computer Science.

## APPENDIX I

*Proof of Theorem 1:* The first assertion is trivial; we only prove the second one. Since $t' = t'_{ij}$, there is a set $\mathfrak{R}_{ij}$ of rules associated to it. Following Step 7) in the algorithm for extending FPNCWs, we can readily obtain a state distribution $D_{t'_{ij}}$ by a direct computation, that is, for any $p \in P' = P$,

$$\begin{aligned}D_{t'_{ij}}(p) &= \bigvee_{t_s \in T(\mathfrak{R}_{ij})} \bigvee_{a \in \Sigma} \left(l(t_s)(a) \wedge l(t'_{ij})(a) \wedge D_{t_s}(p)\right) \\ &= \bigvee_{t_s \in T(\mathfrak{R}_{ij})} \bigvee_{a \in \Sigma} \left(l(t_s)(a) \wedge l(t'_{ij})(a) \wedge \beta(t_s, p)\right) \\ &= \bigvee_{t_s \in T(\mathfrak{R}_{ij})} \left[\text{height}\left(l(t_s) \cap l(t'_{ij})\right) \wedge \beta(t_s, p)\right].\end{aligned}$$

To check the assertion 2), three cases need to be considered:
Case 1: $p \in I'(t')$. In this case, we also see that $p \in I(t_s)$ for any $t_s \in T(\mathfrak{R}_{ij})$. Therefore, by definition,

$$f(M, t_s)(p) = M(p) \vee (\mu_{M,t_s} \wedge \beta(t_s, p))$$

for any $t_s \in T(\mathfrak{R}_{ij})$. On the other hand, we have the first equation at the top of the next page, as desired. Note that in the computation, we have used the fact $\mu_{M,t'_{ij}} = \mu_{M,t_s} = \mu_{M,t'_{s'}}$ for any $t_s, t_{s'} \in T(\mathfrak{R}_{ij})$.
Case 2: $p \in I'(t') \bigcap O'(t')$. In this case, for any $t_s \in T(\mathfrak{R}_{ij})$ we still have that $p \in I(t_s)$, but it is possible that $p \notin O(t_s)$. Hence, we set $T_o(\mathfrak{R}_{ij}) = \{t_s \in T(\mathfrak{R}_{ij}) : p \in O(t_s)\}$. It follows from definition that

$$f(M, t_s)(p) = \begin{cases} \mu_{M,t_s} \wedge \beta(t_s, p), & \text{if } t_s \in T_o(\mathfrak{R}_{ij}) \\ 0, & \text{otherwise.} \end{cases}$$

Observe that in this case,

$$D'_{ij}(p) = \bigvee_{t_s \in T_o(\mathfrak{R}_{ij})} \left[\text{height}\left(l(t_s) \cap l(t'_{ij})\right) \wedge \beta(t_s, p)\right].$$

Thus, we have the second equation at the top of this page.
Case 3: $p \in I'(t') \setminus O'(t')$. In this case, $f'(M, t')(p) = 0$ by definition. On the other hand, it is clear that $p \in I(t_s) \setminus O(t_s)$ for any $t_s \in T(\mathfrak{R}_{ij})$, so $f(M, t_s)(p) = 0$ for all $t_s \in T(\mathfrak{R}_{ij})$. Consequently, $f'(M, t')(p)$ can be written in the form $\bigvee_{t_s \in T(\mathfrak{R}_{ij})} \left[\text{height}\left(l(t_s) \cap l(t'_{ij})\right) \wedge f(M, t_s)(p)\right]$.
This completes the proof of the theorem.

*Proof of Proposition 1:* To prove $L_{\widetilde{\mathcal{N}}_M} = L_M$, we need to verify that $L_{\widetilde{\mathcal{N}}_M}(S) = L_M(S)$ for any $S \in \widetilde{\Sigma}^*$. By definition, $L_{\widetilde{\mathcal{N}}_M}(S) = \text{height}\left[\left(\bigcup_{t \in T_S^*} f(M_0, t)\right) \cap M_1\right]$ and $L_M(S) = \text{height}(\delta(q_0, S) \cap F)$. Therefore, it suffices to show that

$$\left(\bigcup_{t \in T_S^*} f(M_0, t)\right)(q) = \delta(q_0, S)(q) \tag{1}$$

for any $q \in Q = P$. Assume that $S = W_0 W_1 \cdots W_k$ for some $W_i \in \widetilde{\Sigma}$, $i = 0, 1, \ldots, k$. Note that $\left(\bigcup_{t \in T_S^*} f(M_0, t)\right)(q) = \bigvee_{t \in T_S^*} f(M_0, t)(q)$ and $\delta(q_0, S)(q) = \vee\{\delta(q_0, W_0)(q_1) \wedge \delta(q_1, W_1)(q_2) \wedge \cdots \wedge \delta(q_k, W_k)(q_{k+1}) : W_0 W_1 \cdots W_k = S, q_1, q_2, \ldots, q_k \in Q, q_{k+1} = q\}$. Hence, to prove (1), we only need to verify the following:

(i) For any trace $q_0 W_0 q_1 W_1 q_2 \cdots q_k W_k q$ in $M$ with $\delta(q_0, W_0)(q_1) \wedge \delta(q_1, W_1)(q_2) \wedge \cdots \wedge \delta(q_k, W_k)(q) > 0$, there is a $t \in T_S^*$ such that $f(M_0, t)(q) = \delta(q_0, W_0)(q_1) \wedge \delta(q_1, W_1)(q_2) \wedge \cdots \wedge \delta(q_k, W_k)(q)$.

$$\begin{aligned}
f'(M,t')(p) &= M(p) \vee (\mu_{M,t_s} \wedge \beta'(t',p)) \\
&= M(p) \vee (\mu_{M,t_s} \wedge D_{t'_{ij}}(p)) \\
&= M(p) \vee \left\{ \mu_{M,t_s} \wedge \bigvee_{t_s \in T(\mathfrak{R}_{ij})} \left[ \text{height}\left(l(t_s) \cap l(t'_{ij})\right) \wedge \beta(t_s,p) \right] \right\} \\
&= M(p) \vee \left\{ \bigvee_{t_s \in T(\mathfrak{R}_{ij})} \left[ \text{height}\left(l(t_s) \cap l(t'_{ij})\right) \wedge \mu_{M,t_s} \wedge \beta(t_s,p) \right] \right\} \\
&= \bigvee_{t_s \in T(\mathfrak{R}_{ij})} \left\{ \left[ M(p) \vee \text{height}\left(l(t_s) \cap l(t'_{ij})\right) \right] \wedge \left[ M(p) \vee (\mu_{M,t_s} \wedge \beta(t_s,p)) \right] \right\} \\
&= \bigvee_{t_s \in T(\mathfrak{R}_{ij})} \left\{ \left[ M(p) \vee \text{height}\left(l(t_s) \cap l(t'_{ij})\right) \right] \wedge f(M,t_s)(p) \right\} \\
&= \bigvee_{t_s \in T(\mathfrak{R}_{ij})} \left\{ \left[ M(p) \vee \text{height}\left(l(t_s) \cap l(t'_{ij})\right) \right] \wedge \left[ M(p) \vee f(M,t_s)(p) \right] \right\} \\
&= \bigvee_{t_s \in T(\mathfrak{R}_{ij})} \left\{ M(p) \vee \left[ \text{height}\left(l(t_s) \cap l(t'_{ij})\right) \wedge f(M,t_s)(p) \right] \right\} \\
&= M(p) \vee \bigvee_{t_s \in T(\mathfrak{R}_{ij})} \left[ \text{height}\left(l(t_s) \cap l(t'_{ij})\right) \wedge f(M,t_s)(p) \right].
\end{aligned}$$

$$\begin{aligned}
f'(M,t')(p) &= \mu_{M,t_s} \wedge \beta'(t',p) \\
&= \mu_{M,t_s} \wedge D'_{ij}(p) \\
&= \mu_{M,t_s} \wedge \bigvee_{t_s \in T_o(\mathfrak{R}_{ij})} \left[ \text{height}\left(l(t_s) \cap l(t'_{ij})\right) \wedge \beta(t_s,p) \right] \\
&= \bigvee_{t_s \in T_o(\mathfrak{R}_{ij})} \left[ \text{height}\left(l(t_s) \cap l(t'_{ij})\right) \wedge \mu_{M,t_s} \wedge \beta(t_s,p) \right] \\
&= \bigvee_{t_s \in T_o(\mathfrak{R}_{ij})} \left[ \text{height}\left(l(t_s) \cap l(t'_{ij})\right) \wedge f(M,t_s)(p) \right] \\
&= \bigvee_{t_s \in T(\mathfrak{R}_{ij})} \left[ \text{height}\left(l(t_s) \cap l(t'_{ij})\right) \wedge f(M,t_s)(p) \right].
\end{aligned}$$

(ii) For any $t \in T_S^*$ with $f(M_0,t)(q) > 0$, there is a trace $q_0 W_0 q_1 W_1 q_2 \cdots q_k W_k q$ in $M$ such that $\delta(q_0,W_0)(q_1) \wedge \delta(q_1,W_1)(q_2) \wedge \cdots \wedge \delta(q_k,W_k)(q) = f(M_0,t)(q)$.

For (i), let $t_i = t_{(q_i,W_i,q_{i+1})}$, $i = 0,1,\ldots,k$. Then it is easy to see that every $t_i$ is a transition of $\widetilde{\mathcal{N}}_M$. Taking $t = t_0 t_1 \cdots t_k$, we get that $l(t) = S$ and thus $t \in T_S^*$. Furthermore, it follows that $f(M_0,t)(q) = \delta(q_0,W_0)(q_1) \wedge \delta(q_1,W_1)(q_2) \wedge \cdots \wedge \delta(q_k,W_k)(q)$ by the construction and the definition of state transition function. Conversely, for (ii), suppose that $t = t'_0 t'_1 \cdots t'_k \in T_S^*$. Then each $t'_i$ is of the form $t_{(q'_i,W_i,q'_{i+1})}$ for some $q'_i, q'_{i+1} \in Q$. Since $f(M_0,t)(q) > 0$, it forces that $q'_0 = q_0$ and $q'_{k+1} = q$. We thus obtain a trace $q_0 W_0 q'_1 W_1 q'_2 \cdots q'_k W_k q$ in $M$. By the construction, this trace gives that $\delta(q_0,W_0)(q'_1) \wedge \delta(q'_1,W_1)(q'_2) \wedge \cdots \wedge \delta(q'_k,W_k)(q) = f(M_0,t)(q)$. This completes the proof of the proposition.

*Proof of Proposition 2:* For any $S \in \widetilde{\Sigma}^*$, we have by definition and the construction that

$$\begin{aligned}
L_{M_{\widetilde{\mathcal{N}}}}(S) &= \text{height}[\delta(q_0,S) \cap F] \\
&= \bigvee_{q' \in Q} [\delta(q_0,S)(q') \wedge F(q')] \\
&= \bigvee_{t \in T_S^*} \text{height}[f(M_0,t) \cap M_1] \\
&= \text{height}\left[ \left( \bigcup_{t \in T_S^*} f(M_0,t) \right) \cap M_1 \right] \\
&= L_{\widetilde{\mathcal{N}}}(S).
\end{aligned}$$

Consequently, $L_{M_{\widetilde{\mathcal{N}}}} = L_{\widetilde{\mathcal{N}}}$, as desired.

## REFERENCES

[1] A. J. Bugarin and S. Barro, "Fuzzy reasoning supported by Petri nets," *IEEE Trans. Fuzzy Syst.*, vol. 2, pp. 135-150, 1994.
[2] T. Cao and A. C. Sanderson, "Task sequence planning using fuzzy Petri nets," *IEEE Trans. Syst., Man, Cybern.*, vol. 25, pp. 755-768, May 1995.
[3] Y. Cao, L. Xia, and M. Ying, "Probabilistic automata for computing with words," available at *http://arxiv.org/abs/cs/0604087v1*, Apr. 2006.
[4] Y. Cao and M. Ying, "Supervisory control of fuzzy discrete event systems," *IEEE Trans. Syst., Man, Cybern., Part B*, vol. 35, pp. 366-371, Apr. 2005.
[5] ——, "Observability and decentralized control of fuzzy discrete event systems," *IEEE Trans. Fuzzy Syst.*, vol. 14, pp. 202-216, Apr. 2006.
[6] Y. Cao, M. Ying, and G. Chen, "Retraction and generalized extension of computing with words," *IEEE Trans. Fuzzy Syst.*, vol. 15, pp. 1238-1250, Dec. 2007.

[7] J. Cardoso, R. Valette, and D. Dubois, "Fuzzy Petri net: an overview," in *Proc. 13th IFAC World Congr.*, San Francisco, CA, June 30-July 5, 1996, pp. 443-448.

[8] S. M. Chen, J. S. Ke, and J. F. Chang, "Knowledge representation using fuzzy Petri nets," *IEEE Trans. Knowl. Data Eng.*, vol. 2, no. 3, pp. 311-319, Sep. 1990.

[9] M. Gao, M. C. Zhou, X. Huang, and Z. Wu, "Fuzzy reasoning Petri nets," *IEEE Trans. Syst., Man, Cybern., Part A*, vol. 33, no. 3, pp. 314-324, May 2003.

[10] X. G. He, "Fuzzy Petri net," *Chinese J. Comput.*, vol. 17, no. 12, pp. 946-950, 1994, (in Chinese).

[11] F. Herrera and L. Martínez, "A 2-tuple fuzzy linguistic representation model for computing with words," *IEEE Trans. Fuzzy Syst.*, vol. 8, pp. 746-752, Dec. 2000.

[12] R. I. John and P. R. Innocent, "Modeling uncertainty in clinical diagnosis using fuzzy logic," *IEEE Trans. Syst., Man, Cybern., Part B*, vol. 35, pp. 1340-1350, Dec. 2005.

[13] G. J. Klir and B. Yuan, *Fuzzy Sets and Fuzzy Logic: Theory and Applications.* Upper Saddle River, NJ: Prentice-Hall, 1995.

[14] L. T. Kóczy and K. Hirota, "Size reduction by interpolation in fuzzy rule bases," *IEEE Trans. Syst., Man Cybern., Part B*, vol. 27, no. 1, pp. 14-25, Feb. 1997.

[15] J. Lee, K. F. R. Liu, and W. Chiang, "Modeling uncertainty reasoning with possibilistic Petri nets," *IEEE Trans. Syst., Man Cybern., Part B*, vol. 33, no. 2, pp. 214-224, Apr. 2003.

[16] F. Lin and H. Ying, "Modeling and control of fuzzy discrete event systems," *IEEE Trans. Syst., Man, Cybern., Part B*, vol. 32, pp. 408-415, Aug. 2002.

[17] T. Y. Lin, "Granular computing: From rough sets and neighborhood systems to information granulation and computing in words," in *Eur. Congr. Intell. Tech. Soft Comput.*, Sept. 8-12, 1997, pp. 1602-1606.

[18] T. Y. Lin, "Granular computing on binary relations I: Data mining and neighborhood systems," in *Rough Sets In Knowledge Discovery*, A. Skowron and L. Polkowski (eds), Physica-Verlag, pp. 107-121, 1998.

[19] T. Y. Lin, "Granular computing on binary relations II: Rough set representations and belief functions," in *Rough Sets In Knowledge Discovery*, A. Skowron and L. Polkowski (eds), Physica-Verlag, pp. 121-140, 1998.

[20] C. G. Looney, "Fuzzy Petri nets for rule-based decisionmaking," *IEEE Trans. Syst., Man, Cybern.*, vol. 18, no. 1, pp. 178-183, Jan./Feb. 1988.

[21] E. H. Mamdani, "Application of fuzzy logic to approximate reasoning using linguistic synthesis," *IEEE Trans. Comput.*, vol. 26, pp. 1182-1191, Dec. 1977.

[22] M. Margaliot and G. Langholz, "Fuzzy control of a benchmark problem: A computing with words approach," *IEEE Trans. Fuzzy Syst.*, vol. 12, pp. 230-235, Apr. 2004.

[23] J. M. Mendel, "Computing with words, when words can mean different things to different people," in *Proc. 3rd Int'l ICSC Symp. Fuzzy Logic Appl.*, Rochester, NY, June 1999, pp. 158-164.

[24] ——, "The perceptual computer: An architecture for computing with words," in *Proc. FUZZ-IEEE*, Melbourne, Australia, Dec. 2001, pp. 35-38.

[25] ——, "Computing with words and its relationships with fuzzistics," *Inform. Sci.*, vol. 177, pp. 988-1006, 2007.

[26] J. M. Mendel and D. Wu, "Perceptual reasoning for perceptual computing," *IEEE Trans. Fuzzy Syst.*, vol. 16, no. 6, pp. 1550-1564, Dec. 2008.

[27] R. Milner, *Communicating and Mobile Systems: The $\pi$-Calculus.* Cambridge: Cambridge University Press, 1999.

[28] T. Murata, "Petri nets: Properties, analysis and applications," *Proc. IEEE*, vol. 77, pp. 541-580, Apr. 1989.

[29] W. Pedrycz and F. Gomide, "A generalized fuzzy Petri net model," *IEEE Trans. Fuzzy Syst.*, vol. 2, no. 4, pp. 295-301, Nov. 1994.

[30] W. Pedrycz and F. Gomide, *An Introduction to Fuzzy Sets: Analysis and Design.* Cambridge, Mass: MIT Press, 1998.

[31] J. L. Peterson, *Petri Net Theory and the Modeling of Systems.* Englewood Cliffs, NJ: Prentice-Hall, 1981.

[32] C. A. Petri, *Kommunikation mit Automaten.* Ph.D. dissertation, Bonn: University of Bonn, 1962, (in German).

[33] D. W. Qiu, "Supervisory control of fuzzy discrete event systems: A formal approach," *IEEE Trans. Syst., Man, Cybern., Part B*, vol. 35, pp. 72-88, Feb. 2005.

[34] D. W. Qiu and H. Q. Wang, "A probabilistic model of computing with words," *J. Comput. Syst. Sci.*, vol. 70, pp. 176-200, 2005.

[35] D. Sangiorgi, "On the origins of bisimulation and coinduction," *ACM Trans. Progr. Lang. Syst.*, vol. 31, pp. 111-151, 2009.

[36] V. R. L. Shen, "Knowledge representation using high-level fuzzy petri nets," *IEEE Trans. Syst., Man, Cybern., Part A: Syst. Humans*, vol. 36, pp. 1220-1227, Nov. 2006.

[37] W. Siler and J. J. Buckley, *Fuzzy Expert Systems and Fuzzy Reasoning.* Hoboken, New Jersey: John Wiley & Sons, Inc., 2005.

[38] F. Y. Wang, "On the abstraction of conventional dynamic systems: From numerical analysis to linguistic analysis," *Inform. Sci.*, vol. 171, pp. 233-259, 2005.

[39] H. Q. Wang and D. W. Qiu, "Computing with words via Turing machines: A formal approach," *IEEE Trans. Fuzzy Syst.*, vol. 11, pp. 742-753, Dec. 2003.

[40] J. H. Wang and J. Hao, "A new version of 2-tuple fuzzy linguistic representation model for computing with words," *IEEE Trans. Fuzzy Syst.*, vol. 14, no. 3, pp. 435-445, Jun. 2006.

[41] J. H. Wang and J. Hao, "An approach to computing with words based on canonical characteristic values of linguistic labels," *IEEE Trans. Fuzzy Syst.*, vol. 15, no. 4, pp. 593-604, Aug. 2007.

[42] R. R. Yager, "Defending against strategic manipulation in uninorm-based multi-agent decision making," *Eur. J. Oper. Res.*, vol. 141, pp. 217-232, Aug. 2002.

[43] ——, "On the retranslation process in Zadeh's paradigm of computing with words," *IEEE Trans. Syst., Man, Cybern., Part B*, vol. 34, pp. 1184-1195, Apr. 2004.

[44] M. S. Ying, "A formal model of computing with words," *IEEE Trans. Fuzzy Syst.*, vol. 10, pp. 640-652, Oct. 2002.

[45] L. A. Zadeh, "Fuzzy sets," *Inform. Contr.*, vol. 8, pp. 338-353, 1965.

[46] ——, "Fuzzy Logic = computing with words," *IEEE Trans. Fuzzy Syst.*, vol. 4, pp. 103-111, Apr. 1996.

[47] ——, "From computing with numbers to computing with words – From manipulation of measurements to manipulation of perceptions," *IEEE Trans. Circuits Syst. I: Fund. Theory Appl.*, vol. 45, pp. 105-119, Jan. 1999.

[48] ——, "Outline of a computational theory of perceptions based on computing with words," in *Soft Computing and Intelligent Systems*, N. K. Sinha and M. M. Gupta, Eds. Boston, MA: Academic, 1999, pp. 3-22.

[49] ——, "A new direction in AI: Toward a computational theory of perceptions," *AI Mag.*, vol. 22, pp. 73-84, 2001.

[50] ——, "Toward a perception-based theory of probabilistic reasoning with imprecise probabilities," *J. Stat. Plan. Infer.* vol. 105, pp. 233-264, 2002.

[51] L. A. Zadeh and J. Kacprzyk, *Computing with Words in Information/Intelligent Systems.* Heidelberg, Germany: Physica-Verlag, vol. 1 and vol. 2, 1999.